\documentclass[letterpaper, 10 pt, conference]{ieeeconf}  % Comment this line out if you need a4paper

\IEEEoverridecommandlockouts                              % This command is only needed if 
\usepackage{graphics} % for pdf, bitmapped graphics files
\usepackage{epsfig} % for postscript graphics files
\usepackage[noend]{algpseudocode}
\usepackage{algorithm}
\usepackage[ruled,vlined,algo2e]{algorithm2e}
\providecommand{\SetAlgoLined}{\SetLine}
\usepackage{multirow}
\usepackage{mathptmx} % assumes new font selection scheme installed
\usepackage{times} % assumes new font selection scheme installed
\usepackage{amsmath} % assumes amsmath package installed
\usepackage{amssymb}  % assumes amsmath package installed
\usepackage{dsfont}	
\usepackage{bbm}
\newcommand{\bfsection}[1]{\vspace*{0.1cm}\noindent\textbf{#1.}}
\usepackage{footnote}
\title{\LARGE \bf
FIDNet: LiDAR Point Cloud Semantic Segmentation with Fully Interpolation Decoding} 
\author{Yiming Zhao, Lin Bai and Xinming Huang% <-this % stops a space
\thanks{Authors are with Department of Electrical and Computer Engineering, Worcester Polytechnic Institute,
        Massachusetts 01609, USA.
        yzhao7@wpi.edu}%
}

\begin{document}

\maketitle
\thispagestyle{empty}
\pagestyle{empty}

%%%%%%%%%%%%%%%%%%%%%%%%%%%%%%%%%%%%%%%%%%%%%%%%%%%%%%%%%%%%%%%%%%%%%%%%%%%%%%%%

\textbf{Note: }  After the submission of IROS, we have some follow-up updates, please scroll to the end and take a look at the \textbf{supplementary file} after the reference section. Further details can be found in our code: https://github.com/placeforyiming/IROS21-FIDNet-SemanticKITTI \newline

\begin{abstract}

Projecting the point cloud on the 2D spherical range image transforms the LiDAR semantic segmentation to a 2D segmentation task on the range image. However, the LiDAR range image is still naturally different from the regular 2D RGB image; for example, each position on the range image encodes the unique geometry information. In this paper, we propose a new projection-based LiDAR semantic segmentation pipeline that consists of a novel network structure and an efficient post-processing step. In our network structure, we design a FID (fully interpolation decoding) module that directly upsamples the multi-resolution feature maps using bilinear interpolation. Inspired by the 3D distance interpolation used in PointNet++, we argue this FID module is a 2D version distance interpolation on $(\theta, \phi)$ space. As a parameter-free decoding module, the FID largely reduces the model complexity by maintaining good performance. Besides the network structure, we empirically find that our model predictions have clear boundaries between different semantic classes. This makes us rethink whether the widely used K-nearest-neighbor post-processing is still necessary for our pipeline. Then, we realize the many-to-one mapping causes the blurring effect that some points are mapped into the same pixel and share the same label. Therefore, we propose to process those occluded points by assigning the nearest predicted label to them. This NLA (nearest label assignment) post-processing step shows a better performance than KNN with faster inference speed in the ablation study. On SemanticKITTI dataset, our pipeline achieves the best performance among all projection-based methods with $64 \times 2048$ resolution and all point-wise solutions. With a ResNet-34 as the backbone, both the training and testing of our model can be finished on a single RTX 2080 Ti with 11G memory. The code is released here.\footnote{https://github.com/placeforyiming/IROS21-FIDNet-SemanticKITTI \hspace{0.2cm}}

\end{abstract}

%%%%%%%%%%%%%%%%%%%%%%%%%%%%%%%%%%%%%%%%%%%%%%%%%%%%%%%%%%%%%%%%%%%%%%%%%%%%%%%%
\section{INTRODUCTION}

\begin{figure}
\includegraphics[width=1.0\linewidth,height=50mm]{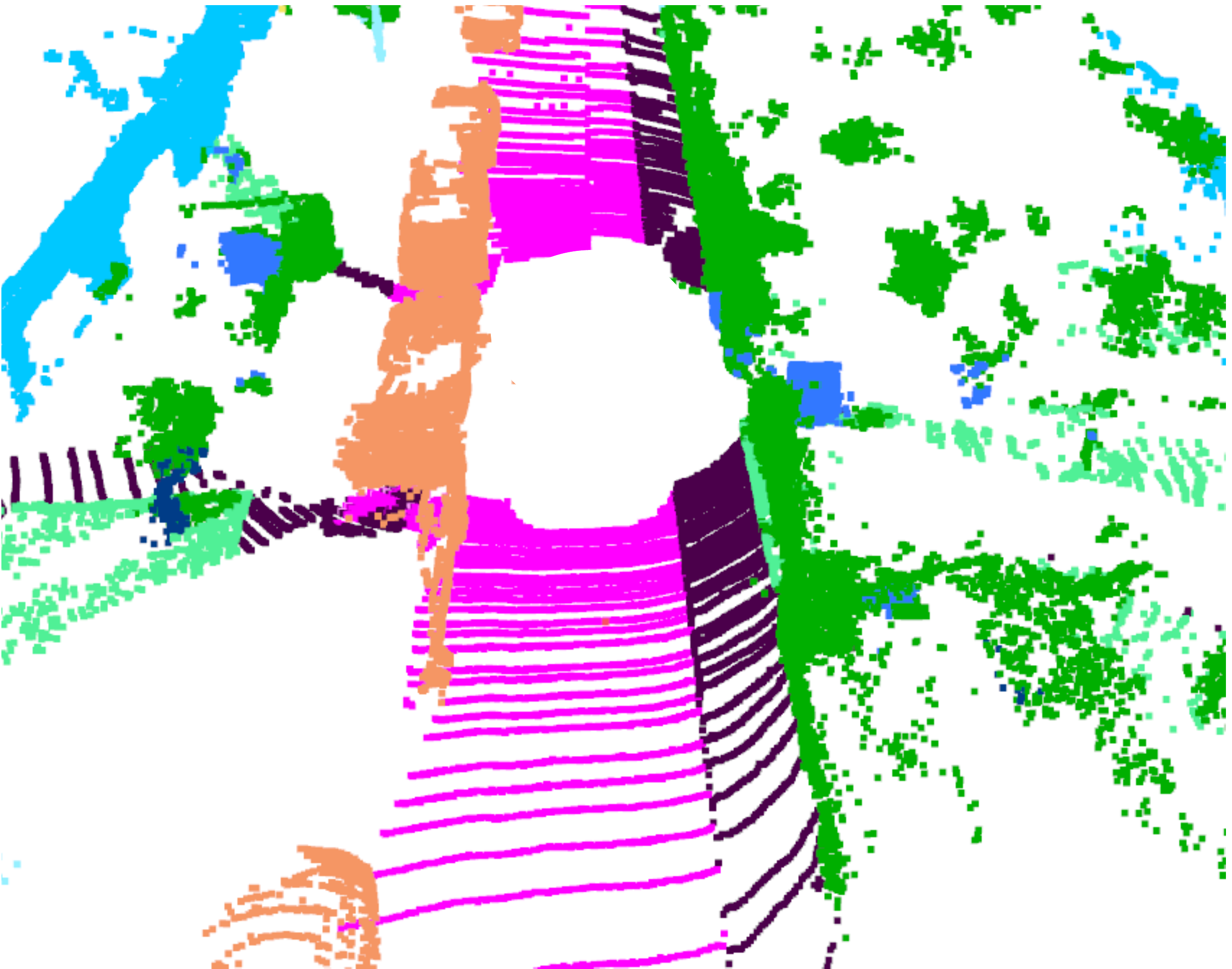}
    \caption{The LiDAR point cloud semantic segmentation predicts a semantic label for each point to help the car to understand the 3D surroundings. This is a sample from the validation sequence in the SemanticKITTI dataset.}
    \label{fig:fisrt}
    \vspace{-6mm}
\end{figure}

LiDAR sensor is playing an important role in outdoor robots, especially autonomous cars. With the booming of deep learning techniques, recent research topics focus on extracting object and semantic information from the point cloud. LiDAR semantic segmentation is such a task that a neural network predicts the semantic label of each point \cite{milioto2019rangenet++}. As shown in Fig. \ref{fig:fisrt}, solving this task builds the 3D understanding of the nearby environment.

Though the first dataset specifically aimed at benchmarking the LiDAR semantic segmentation was published in 2019 \cite{behley2019semantickitti}, the interest in processing 3D point clouds with neural networks arose earlier in two communities. Researchers from the computer vision society investigated how to design a permutation invariant network to deal with more general unordered point clouds \cite{qi2017pointnet,qi2017pointnet++}. In contrast, the solution design from researchers on the robotic side considered more about the LiDAR sensing mechanism by projecting the point cloud on the 2D spherical range image \cite{wu2018squeezeseg}. The projection-based solution allows the well-designed 2D image semantic segmentation models to be directly used for the LiDAR semantic segmentation task. It lacks the ability to process more general unordered point clouds, but it shows practical advantages such as better performance in terms of both the speed and accuracy \cite{hu2020randla, cortinhal2020salsanext}. To chase a better performance, recent researchers further design models by combining multi-view projections or voxelization with point-wise features \cite{cheng20212, zhang2020deep, alnaggar2021multi}.

\textbf{Motivation} The major goal of this paper is to argue how should we design the network for projection-based LiDAR segmentation. We hope to keep the structure as common as possible as well as maintain a good performance. Most image-based network structures rely on an encoder-decoder structure. This paper rethinks this structure by considering the difference between the spherical representation image and the regular image. On the spherical representation image, each position is a 3D point with its unique location information. However, on the regular image, two different pixels may have the same spectral information that makes only a pattern of pixels meaningful. To design a better structure for range image, we propose several modules, including an input module with $1 \times 1$ convolution, a backbone network to extract multi-scale features, a fully interpolation decoding module, and a final post-processing step that only assigning labels for those occluded points.   

\textbf{Contribution} This paper proposes a new pipeline to solve the LiDAR semantic segmentation in a projection fashion. The whole solution is clean and effective. Both the training and testing can be conducted on a single RTX 2080 Ti with 11 G memory. The solution's performance is better than all projection-based methods by following the same $64 \times 2048$ input resolution. There are two major technical contributions that we think will be helpful for other related methods:

\begin{itemize}
    \item \emph{We propose a parameter-free fully interpolation decoding module that only contains bilinear interpolation operation.} Most existing decoder structures need transpose convolution or special combined convolution with interpolation to upsample low-resolution feature maps. We demonstrate that only using bilinear interpolation can already achieve state-of-the-art performance. This setting reduces the model complexity by avoiding a large amount parameters in the decoder used by other models \cite{cheng2020panoptic, cortinhal2020salsanext}. 
    
    \item \emph{We replace the widely used K-nearest-neighbor post-processing with a more efficient and intuitive step.} K-nearest-neighbor was proposed to solve the boundary-blurring effect generated by two reasons \cite{milioto2019rangenet++}, which are the ambiguous output of CNNs and the many-to-one mapping on spherical range image. However, we empirically find the blurring effect on our predictions is negligible. Then, we specifically focus on the many-to-one mapping that some points will be mapped on the same location with others, thus not having directly predicted labels from the network. To solve this problem, we assign the predicted label of the nearest point in 3D space to those unpredicted points. Compared with KNN, this NLA (nearest label assignment) post-processing step has better performance and faster inference speed.
\end{itemize}

Both of the above two contributions bring practical improvements to the network design of projection-based LiDAR semantic segmentation task, which have not been discussed by other literature as far as we know. There are also some other minor contributions such as an input module with $1 \times 1$ convolution or nearby point feature aggregation with atrous convolution \cite{chen2017rethinking}. Since similar ideas have been explored in recent papers \cite{alonso20203d}, we will introduce them in the method part, but will not claim them as the major contribution of this paper. 

\section{RELATED WORK}

\subsection{Point-based Networks} 
\textbf{PointNets} PointNet \cite{qi2017pointnet} summarizes several key properties of the general point cloud, including unordered, invariance under transformations, and interaction among points. Those unique challenges stimulate them to design a MLP (multi-layer perceptron) based network structure and a per-point feature vector concatenated by global and point-wise features. Besides only the concatenation of global and point-wise features, PointNet++ \cite{qi2017pointnet++}  propose novel set learning layers to adaptively combine features from multiple scales. However, the local query and grouping limit the model performance on a large point cloud. This attracts some successive papers \cite{landrieu2018large, hu2020randla}, the best pure point-based solution still lags behind on LiDAR point cloud semantic segmentation \cite{hu2020randla}.

\textbf{PointConvs} Along with the network structure designing, solving the point cloud segmentation with special convolutional kernels is also a hot topic. Some ideas, such as PointCNN \cite{li2018pointcnn} and KPConv \cite{thomas2019kpconv}, have been tried on various datasets and show a strong generality. Some of those special convolutional kernels have been merged as part of other solutions \cite{kochanov2020kprnet}.   

\subsection{Voxel-based Networks}
Split the 3D world in discretized voxels is a straightforward idea to process the point cloud. After the voxelization, the regular 3D convolution is able to be used on 3D space just like 2D convolution on 2D image \cite{graham20183d, han2020occuseg}. However, the improvement of those methods in the outdoor LiDAR point cloud remains limited. Some recent papers are trying to consider a better 3D voxelization method to cut the 3D space by incorporating how LiDAR sensor generate the point cloud \cite{zhu2020cylindrical}. 

\subsection{Projection-based Network}
\textbf{Single-View projection}
The success of 2D convolution neural networks on the image raises the question of whether it is possible to project the 3D point cloud on 2D space for the sake of using well-developed existing models. The scanning mechanism of the LiDAR sensor suggests the spherical range projection \cite{milioto2019rangenet++}. This idea has been explored by many recent papers from different aspects \cite{cortinhal2020salsanext, xu2020squeezesegv3, alonso20203d}. Besides the range view, bird-eye-view is also considered by recent papers \cite{zhang2020deep}. All those single view projection methods have the benefit of using 2D convolution networks, like the controllable inference speed.

\textbf{Multi-View projection}
With all those newly developed methods above, it is natural to start thinking about how to combine part of different ideas together. In recent multi-view projection solutions \cite{liong2020amvnet,gerdzhev2020tornado}, MLP based feature extractor is used first, then feature tensors from different views are fused together to be processed by one decoder. However, those methods need to take the extra cost to prepare multi-view projection, and it is also not clear how much improvement compared with single view projection.

\begin{figure*}
\includegraphics[width=1.0\linewidth,height=35mm]{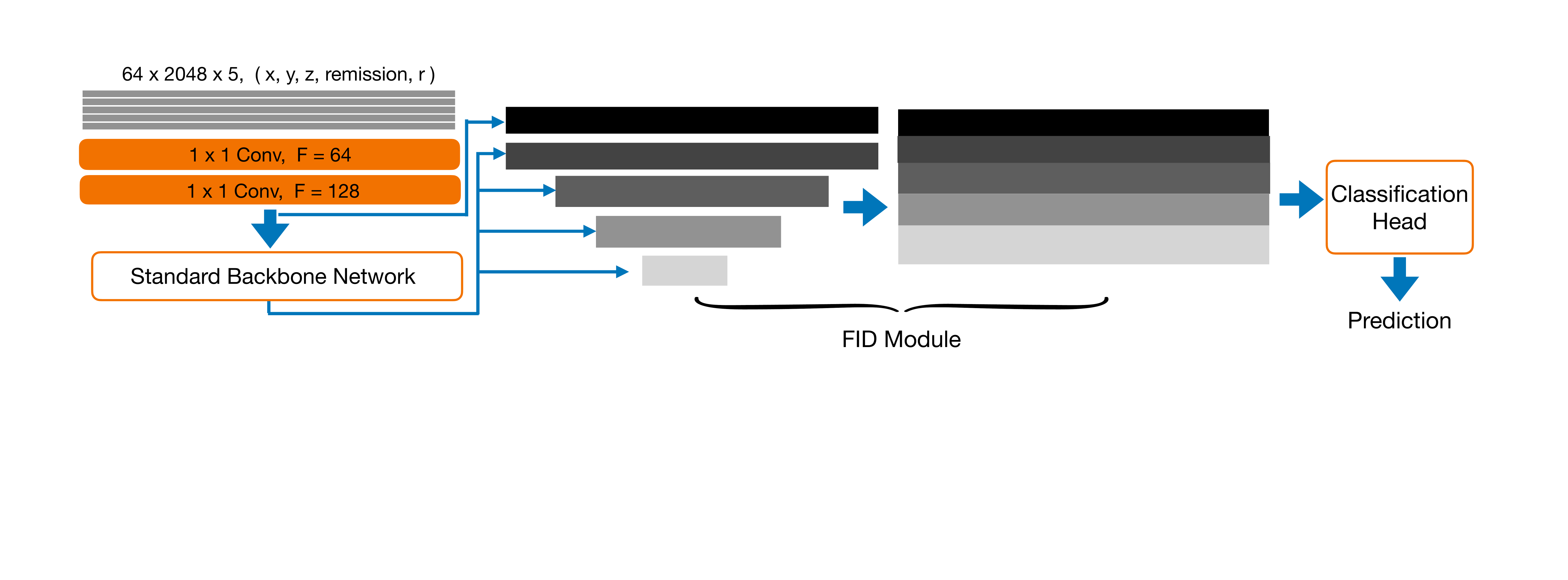}
    \caption{Illustration of our network structure. The input module has two $1 \times 1$ layers mapping each point to a high dimensional space. The backbone can be any regular standard network, like ResNet-34 used in this paper. The FID module upsamples all low-resolution feature maps to the original size and concatenates them together. The last classification head takes in the merged large tensor and outputs the label of each point.   }
    \label{fig:second}
    \vspace{-5mm}
\end{figure*}

\subsection{Practical Considerations}

As a new and important challenge, how to design models to solve the LiDAR semantic segmentation is still an open problem. The solution should consider both the numerical indicator and the practical feasibility. Single view projection methods are able to directly use 2D convolutional neural networks, thus do not need to worry about feasibility as there are a bunch of techniques to optimize 2D networks for various application requirements \cite{xu2018deep,molchanov2019importance}. This paper specifically works on providing a pipeline for spherical range single view projection solutions with better performance. 

\section{METHOD}

The range view projection methods map each point from $(x, y, z)$ Cartesian coordinate to $(r, \theta, \phi)$ spherical coordinate. After discretizing the 2D $(\theta, \phi)$ space, each point will have a mapped position on the $(\theta, \phi)$ image. The same as RGB three channels on regular images, the LiDAR point cloud range image has five channels $(x, y, z, r, remission)$.

\subsection{Input and Backbone Modules}
Though the input is like a 2D image with five channels, each position still represents the information of a point. Therefore, we process the input with two $1 \times 1$ convolutional layers that map each point to a high dimension tensor. This input processing module is analogous to the PointNet \cite{qi2017pointnet} which extracts point features by using MLP (multi-layer perceptron). Then, we feed the high dimensional tensor consists of point-wise features into a regular backbone network. The backbone network can be any structures, such as faster MobileNet \cite{howard2019searching} or more accurate HRNet \cite{wang2020deep}. All those CNN structures take a tensor in and generate multi-resolution feature maps. In this paper, we use the standard ResNet-34 \cite{he2016deep} as the backbone.

The question is, why the regular backbone network designed for images still can work on point-wise feature tensors? We argue that the LiDAR point cloud is ordered in spherical coordinate as the laser scanner emits laser beams line by line. This mechanism makes it possible to build the special 2D spherical range representation of the LiDAR point cloud. The regular $3 \times 3 $ convolutional operator in modern backbone networks is equivalent to the eight nearest queries and grouping on the 2D $(\theta, \phi)$ space. Thus, using the backbone network designed for the image to process point-wise features is the 2D correspondence of the 3D K-nearest query and grouping for unordered point cloud used in PointNet++ \cite{qi2017pointnet++}.

\subsection{Fully Interpolation Decoding}
In our network structure shown in Fig. \ref{fig:second}, we use $1 \times 1$ convolution to map each point vector to a high dimension tensor, then extract multi-scale point features with a backbone network that keeps processing nearby point features on 2D $(\theta, \phi)$ space. The next question is, how to fuse that information together in a point-wise way. In PointNet \cite{qi2017pointnet}, the per-point feature vector is concatenated with the global feature vector as the final feature vector for each point. In PointNet++ \cite{qi2017pointnet++}, the distance-based interpolation is used to upsample a subset of points to a larger set. Those two unique operations inspire us to design a similar module aiming to fuse the multi-scale information.

The distance interpolation defined in PointNet++ \cite{qi2017pointnet++} is:
$$f(x)=\frac{\sum_{i=1}^{k}w_{i}(x)f_{i}}{\sum_{i=1}^{k}w_{i}(x)}, \hspace{2mm} where \hspace{2mm} w_{i}(x)=\frac{1}{d(x,x_{i})}. $$
However, on the 2D $(\theta, \phi)$ space, if we set number of nearest points $k = 4$ and define the distance function $d(x,x_{i})$ as $l_{1}$ distance, the distance interpolation will exactly degenerate to bilinear upsample. Using the bilinear upsample to interpolate the low-resolution feature maps gives us five point-wise feature tensors which have the same resolution but encode different level information. The same as PointNet \cite{qi2017pointnet}, we concatenate all those feature tensors together. Then, after going through the network structure in Fig. \ref{fig:second}, each point is mapped from $(x, y, z, r, remission)$ to a high dimensional vector with multi-level information. 

We name this module FID (fully interpolation decoding). Compared with regular decoder used in other networks, FID is completely parameter-free that largely reduces the model complexity and memory cost. This advantage also brings benefits to the network structure design. For example, one can only focus on customizing the backbone to meet the requirement of the hardware limitation. 

\subsection{Classification Head}

\begin{figure}
 \centering
\includegraphics[width=0.8\linewidth,height=60mm]{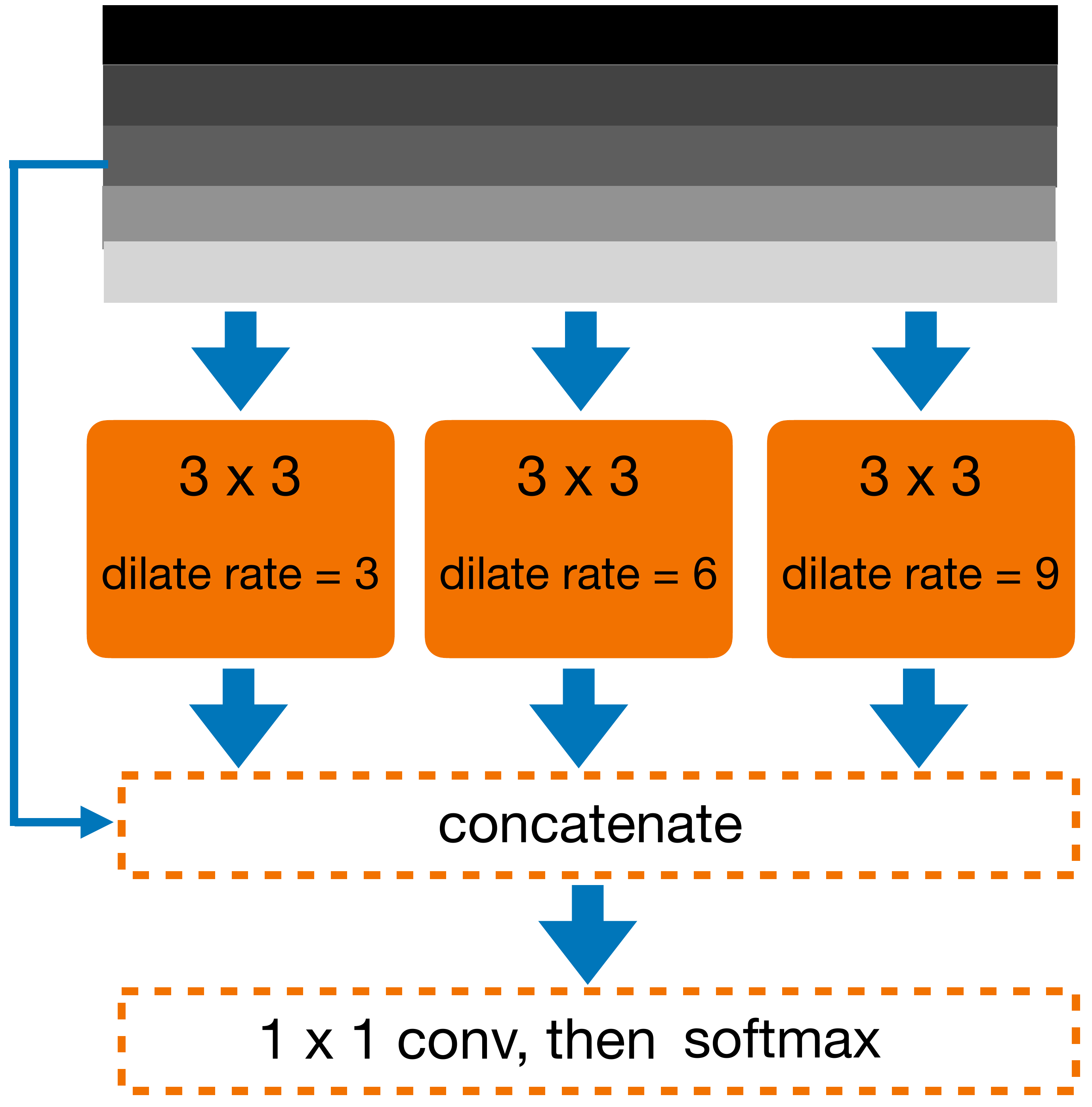}
    \caption{Illustration of our classification head. This head aggregates the original tensor from the FID (fully interpolation decoding) module with tensors from atrous convolution layers. The final point-wise feature tensor is processed by regular $1 \times 1$ convolution layers to get the semantic label.  }
    \label{fig:third}
    \vspace{-3mm}
\end{figure}

After the FID (fully interpolation decoding) module, we get a per-point feature tensor. To help each point vector better fuse the nearby information, we further use atrous convolution to extract local features and aggregate them by simple concatenation with the original feature tensor. This operation is similar to ASPP (atrous spatial pyramid pooling) \cite{chen2017deeplab}, the difference is we do not apply average pooling but concatenate the outputs with original feature tensor together. The final feature tensor is sent into $1 \times 1$ classification layers to predict the semantic label of each point. This classification head is shown in Fig. \ref{fig:third}.

\subsection{Post-processing with Nearest Label Assignment}

\begin{figure}
 \centering
\includegraphics[width=1.0\linewidth,height=45mm]{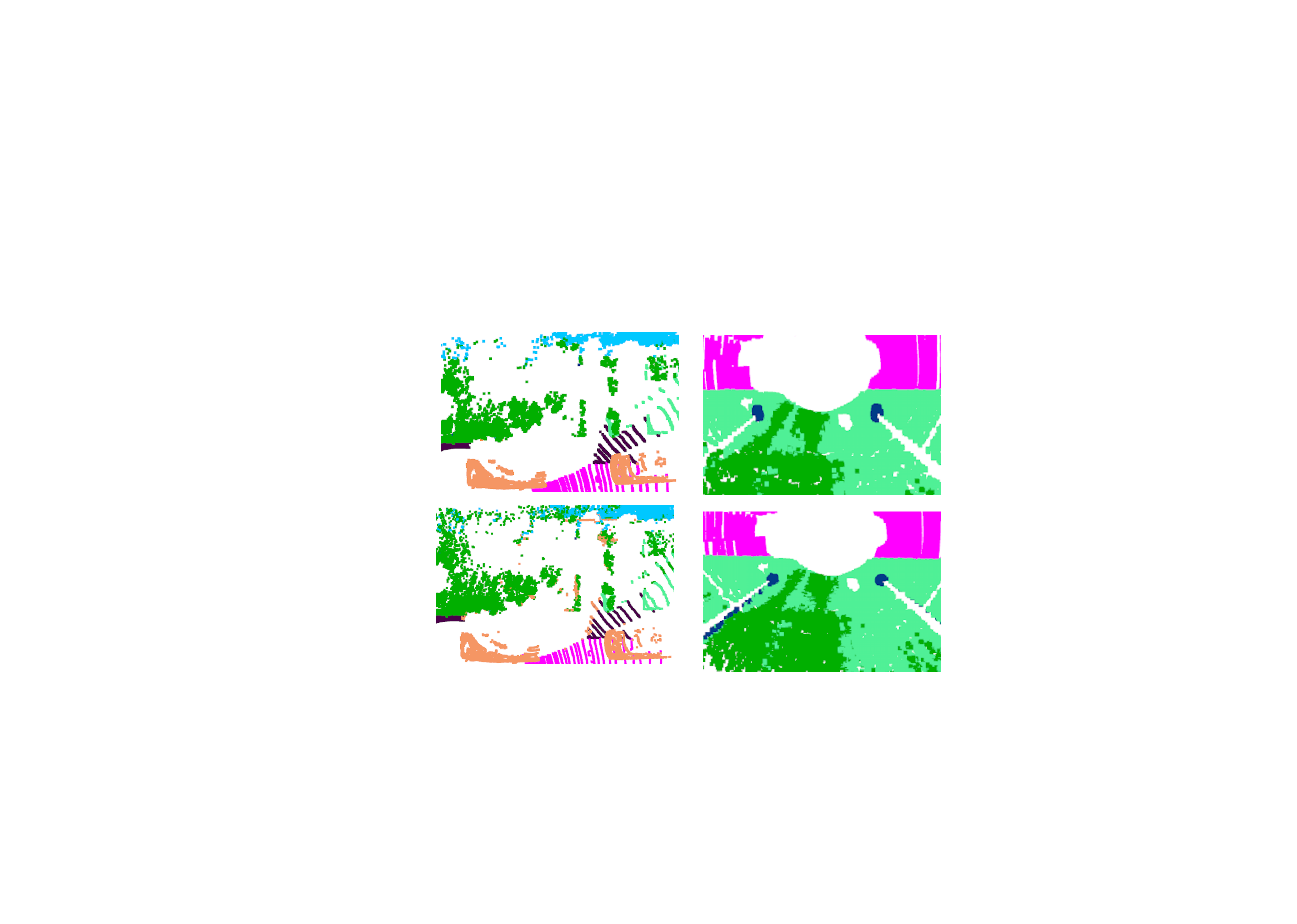}
    \caption{ At the \textbf{(top)}, we visualize those points that are mapped on the range image. Those labels are directly predicted from the neural network and have clear boundaries. At the \textbf{(bottom)}, we add the other points that are mapped on already occupied pixels (more than one point mapped on one pixel). Some of those points are from object boundaries, thus do not share the same label as the predicted one on the pixel. Directly assigning predicted labels will cause the blurred boundary effect as Middle-Red points around the car on the bottom-left image and the Dark-Blue points around poles on the bottom-right image. }
    \label{fig:fourth}
    \vspace{-3mm}
\end{figure}

Almost all recent developed projection-based LiDAR semantic segmentation solutions adopt a KNN (K-nearest-neighbor) post-processing module to alleviate the boundary-blurring effect. As claimed in a recent paper \cite{milioto2019rangenet++}, two reasons make this effect happen: blurring outputs from the neural network and the many-to-one mapping on the range image. However, after visualizing some of our network outputs, we realize the FID (fully interpolation decoding) module is already able to give predictions with clear object boundaries. In Fig. \ref{fig:fourth}, we show our observations. At the top, we visualize two examples by only displaying those points that are processed by the network. We can see there is almost no blurring effect even for small poles. At the bottom of Fig. \ref{fig:fourth}, it is clear to see adding those occluded points creates blurring boundaries. This stimulates us to rethink if it is still necessary to keep the KNN post-processing step in our pipeline.

Here we firstly give more discussions about the many-to-one mapping problem. The mapping from $(x, y, z)$ Cartesian coordinate to $(r, \theta, \phi)$ spherical coordinate is a one-to-one continuous mapping. However, discretizing $(r, \theta, \phi)$ on 2D $(\theta, \phi)$ image will group some close points in one cell. After projection, only the information of one point of the cell will be processed by the neural network. Note, those points mapped on the same cell are only close on 2D $(\theta, \phi)$ space, and they may have a large distance with the processed one. The potential large distance indicates those occluded points may have different labels with the predicted point.

Based on the phenomenon in Fig. \ref{fig:fourth}, we argue the many-to-one mapping is the major issue in our pipeline. From the discussion above, we know those points mapped on the same cell may have large distances with each other, thus do not share the same label. This implies a simple solution that we can directly assign the label of the nearest point in 3D space to those occluded points. Similar KNN, we design our NLA (nearest label assignment), in Alg. \ref{alg:NLA}, on range image to find the nearest point in a local patch that is GPU enabled. 

\begin{algorithm}[H]
	\SetAlgoLined
	\SetKwInOut{Input}{Input}\SetKwInOut{Output}{Output}
    \Input{Range image $I_{r}$ with size $H \times W$,\\
           predicted label map $I_{label}$ with size $H \times W$,\\
           vector $R_{all}$ with range values for all points,\\
           vector $h_{all}$ with projected $h$ values for all points,\\
           vector $w_{all}$ with projected $w$ values for all points,\\
           local kernel size $k$.
           }
	\Output{Vector $Labels$ with predicted labels for all points.}
	\BlankLine
	$Labels \leftarrow \hspace{2mm}empty \hspace{2mm} list \hspace{2mm} [\hspace{2mm}], \hspace{3mm} k \leftarrow 5 \\
	S(h,w,k) \leftarrow \forall (h_{n},w_{m}), \hspace{1mm}where\hspace{1mm} (h_{n},w_{m}) \hspace{1mm}in \hspace{1mm}the \hspace{1mm} k \times k \hspace{1mm} local\hspace{1mm} patch\hspace{1mm} centered \hspace{1mm} at\hspace{1mm} (h,w);$\\ 
	 	%\ForEach{i in range(0, total_number_of_points)}{\\
	    \ForEach{$i$ in $1 : R_{all}.length()$}{
	    $min\_diff \leftarrow +\infty$;
	    \\
	    \ForEach{position $(h_{n},w_{m})$ in $S(h_{all}[i],w_{all}[i],k)$}{
	        \If{$abs(I_{r}(h_{n},w_{m})-R_{all}[i])<min\_diff$}{$label\_each=I_{label}(h_{n},w_{m})\\
	        min\_diff=abs(I_{r}(h_{n},w_{m})-R_{all}[i])$} 
	        
	        $Labels.append(label\_each)$
	    }
	    }
% 	}
	\Return{$Labels$}
	\caption{Nearest Label Assignment (NLA)}\label{alg:NLA}
\end{algorithm}

Compared with KNN, our NLA post-processing step does not need Gaussian weighting and range cutoff, thus is a less complex solution. In the ablation study, we show our nearest label assignment post-processing step has better mIoU with faster inference speed than KNN.

\begin{table*}
\begin{center}
    \caption{The performance comparison on SemanticKITTI test set (sequence 11 to sequence 21). The upper-half are point-wise methods, and the lower-half are projection-based methods.}
        \label{tab:table1}
    \setlength{\tabcolsep}{2.pt}
      \renewcommand{\arraystretch}{1.} 
 \begin{tabular}{c| c| c| c c c c c c c c c c c c c c c c c c c} 
 Methods & Size & \rotatebox{90}{\textbf{mean-IoU}} & \rotatebox{90}{car}& \rotatebox{90}{bicycle}& \rotatebox{90}{motorcycle}& \rotatebox{90}{truck}& \rotatebox{90}{other-vehicle}& \rotatebox{90}{person}& \rotatebox{90}{bicyclist}& \rotatebox{90}{motorcyclist}& \rotatebox{90}{road}& \rotatebox{90}{parking}& \rotatebox{90}{sidewalk}& \rotatebox{90}{other-ground}& \rotatebox{90}{building}& \rotatebox{90}{fence}& \rotatebox{90}{vegetation}& \rotatebox{90}{trunk}& \rotatebox{90}{terrain}& \rotatebox{90}{pole}& \rotatebox{90}{traffic-sign} \\ 
 \hline\hline
 PointNet \cite{qi2017pointnet} & 50K pts &14.6&  46.3 &1.3 &0.3 &0.1& 0.8& 0.2& 0.2& 0.0& 61.6& 15.8& 35.7& 1.4& 41.4& 12.9& 31.0& 4.6& 17.6& 2.4& 3.7\\ 
 PointNet++ \cite{qi2017pointnet++} & 50 K pts&20.1&53.7& 1.9& 0.2& 0.9& 0.2& 0.9& 1.0& 0.0& 72.0& 18.7& 41.8& 5.6& 62.3& 16.9& 46.5& 13.8& 30.0& 6.0 &8.9 \\
 SPGraph \cite{landrieu2018large} & 50K pts &20.0&68.3 &0.9 &4.5& 0.9& 0.8& 1.0& 6.0 &0.0 &49.5& 1.7& 24.2& 0.3& 68.2& 22.5& 59.2& 27.2& 17.0& 18.3& 10.5\\
 SPLATNet \cite{su2018splatnet} & 50K pts &22.8&66.6 &0.0& 0.0& 0.0& 0.0& 0.0& 0.0& 0.0& 70.4& 0.8& 41.5& 0.0& 68.7& 27.8& 72.3& 35.9 &35.8 &13.8 &0.0\\
 TangentConv \cite{tatarchenko2018tangent} & 50K pts & 35.9 &86.8& 1.3& 12.7& 11.6& 10.2& 17.1& 20.2& 0.5 &82.9& 15.2& 61.7& 9.0& 82.8& 44.2& 75.5& 42.5& 55.5& 30.2& 22.2\\
 PointASNL \cite{yan2020pointasnl} & 8K pts &46.8 &87.9& 0.0& 25.1& 39.0& 29.2& 34.2& 57.6& 0.0& 87.4 &24.3& 74.3& 1.8& 83.1& 43.9& 84.1& 52.2& \textbf{70.6}& 57.8& 36.9\\
 LatticeNet \cite{rosu2019latticenet}& 50 K pts &52.9&92.9 &16.6& 22.2& 26.6& 21.4& 35.6& 43.0& \textbf{46.0}& 90.0& 59.4& 74.1& 22.0& 88.2& 58.8& 81.7& 63.6& 63.1& 51.9& 48.4\\
  RandLa-Net \cite{hu2020randla} & 50K pts& 53.9& 94.2& 26.0& 25.8& \textbf{40.1}& 38.9& 49.2 &48.2& 7.2& 90.7& 60.3& 73.7& 20.4& 86.9& 56.3& 81.4& 61.3& 66.8& 49.2& 47.7\\
  S-BKI \cite{gan2020bayesian}& all &51.3& 83.8& 30.6& 43.0& 26.0& 19.6& 8.5& 3.4& 0.0& \textbf{92.6}& \textbf{65.3}& \textbf{77.4}& 30.1& 89.7& 63.7& 83.4& 64.3& 67.4& \textbf{58.6}& \textbf{67.1}\\
  KPConv \cite{thomas2019kpconv} & 50K pts &58.8 &\textbf{96.0}& 30.2& 42.5& 33.4& \textbf{44.3} &61.5& \textbf{61.6}& 11.8& 88.8& 61.3& 72.7& \textbf{31.6}& \textbf{90.5}& 64.2& \textbf{84.8}& \textbf{69.2}& 69.1& 56.4& 47.4\\
 \hline
 
 SqueezeSeg-CRF \cite{wu2018squeezeseg} & $64\times 2048$ &30.8 &68.3 &18.1 &5.1 &4.1 &4.8 &16.5 &17.3& 1.2 &84.9 &28.4 &54.7 &4.6 &61.5& 29.2 &59.6 &25.5 &54.7 &11.2 &36.3 \\ 
 
 SqueezeSegV2-CRF \cite{wu2019squeezesegv2} & $64\times 2048$ &39.6&82.7& 21.0& 22.6& 14.5& 15.9& 20.2& 24.3& 2.9 &88.5& 42.4& 65.5& 18.7& 73.8& 41.0& 68.5& 36.9& 58.9 &12.9& 41.0\\ 
 
 SqueezeSegV3 \cite{xu2020squeezesegv3}& $64\times 2048$&55.9&92.5& 38.7& 36.5& 29.6& 33.0& 45.6& 46.2& 20.1& 91.7& 63.4& 74.8& 26.4& 89.0& 59.4 &82.0& 58.7& 65.4& 49.6& 58.9\\
 
 RangeNet53++KNN \cite{milioto2019rangenet++}& $64\times 2048$&52.2& 91.4 &25.7& 34.4& 25.7& 23.0& 38.3& 38.8& 4.8& 91.8& 65.0& 75.2& 27.8& 87.4& 58.6& 80.5& 55.1& 64.6& 47.9& 55.9\\
 
 SalsaNet \cite{aksoy2019salsanet}& $64\times 2048$&45.4& 87.5& 26.2& 24.6& 24.0& 17.5& 33.2& 31.1& 8.4& 89.7& 51.7& 70.7& 19.7& 82.8& 48.0& 73.0& 40.0& 61.7& 31.3& 41.9\\
 
  SalsaNext \cite{cortinhal2020salsanext}& $64\times 2048$&\textbf{59.5}& 91.9& 48.3& 38.6& 38.9& 31.9& 60.2& 59.0& 19.4& 91.7& 63.7& 75.8& 29.1& 90.2& \textbf{64.2}& 81.8& 63.6& 66.5 &54.3& 62.1 \\

  PolarNet \cite{zhang2020polarnet} &[480, 360, 32] &54.3& 93.8 &40.3& 30.1& 22.9& 28.5& 43.2& 40.2& 5.6& 90.8& 61.7& 74.4& 21.7& 90.0& 61.3& 84.0& 65.5& 67.8& 51.8& 57.5\\
  
  3D-MiniNet-KNN \cite{alonso20203d} & $64\times 2048$&55.8 &90.5& 42.3& 42.1& 28.5& 29.4& 47.8& 44.1& 14.5& 91.6& 64.2& 74.5& 25.4& 89.4& 60.8& 82.8& 60.8& 66.7& 48.0& 56.6\\
  
 \hline
 
 Ours & $64\times 2048$&\textbf{59.5}&93.9&\textbf{54.7}&\textbf{48.9}&27.6&23.9&\textbf{62.3}&59.8&23.7&90.6&59.1&75.8&26.7&88.9&60.5&84.5&64.4&69.0&53.3&62.8\\
 \hline
\end{tabular}
\end{center}
    \vspace{-8mm}
\end{table*}

\subsection{Other Training Settings}

For the data augmentation, we followed other papers to do the rotation and flipping along the y axis \cite{gerdzhev2020tornado, cortinhal2020salsanext}. We set the batch size as 2 and adopted the Adam optimizer with a one-cycle learning rate policy. The maximum learning rate was set to 0.002, and the total training epoch was set to 30. For the loss function, we combined the weighted cross-entropy loss \cite{zhang2018generalized} and the Lovász-Softmax loss \cite{berman2018lovasz} together. Thanks to the parameter-free FID (fully interpolation decoding) module, all our experiments were conducted on a single RTX 2080 Ti with the mix-precision choice in PyTorch.

\section{Experiment}

\subsection{Dataset}

The SemanticKITTI dataset \cite{behley2019semantickitti} is a recent large-scale dataset that provides dense point-wise annotations for the entire KITTI Odometry Benchmark \cite{geiger2012we}. The dataset consists of 22 sequences in total. In this paper, we strictly follow the official split to train the model on sequences 00 to 07, and sequences 09, 10. Sequence 08 is used as the validation set to help us choose the best checkpoint. We submit our predictions on sequences 11 to 21 and report the result from the leaderboard to compare with others.

\subsection{Performance Comparison}

We compare our model performance with other methods in Table \ref{tab:table1}. As we design the structure to process the projected range image by considering point properties, we mainly compare our method with the point-wise solutions as well as the projection-based solutions. By following the same settings, our pipeline outperforms all point-wise solutions with a large margin and achieves similar performance with the best projection-based solutions. 

\subsection{Ablation Study}

In this paper, we empirically find our network predictions do not have a strong edge blurring effect. The blurring on the boundary is mainly caused by the many-to-one mapping problem. To solve this, we design a new post-processing algorithm that simply assigns the label of the nearest point in 3D space to those points without predictions from the neural network. In Table \ref{tab:table2}, we can see this NLA (Nearest Label Assignment) post-processing algorithm achieves better performance with faster inference speed than commonly used KNN. The experiment is conducted on the validation set of SemanticKITTI. The hardware used here is Nvidia RTX 2080 TI. It is also worth saying the CNN structure only needs 11ms to process one sample. Although we use the mix-precision choice provided by PyTorch during training, this inference speed is still remarkable. We credit this benefit to the parameter-free FID module.

\begin{table}
\begin{center}
\caption{Performance comparison between proposed Nearest Label Assignment and commonly used KNN.}
\label{tab:table2}
    \setlength{\tabcolsep}{2.pt}
      \renewcommand{\arraystretch}{1.3} 
\begin{tabular}{ |c|c|c|c|c| } 
\hline
resolution & Network Modules & mIoU (\%) & process time(ms)\\
\hline
\multirow{3}{5em}{$64 \times 2048$} & CNN + & 55.4 & 11  \\ 
\cline{2-4}
&  K-Nearest-Neighbor & 58.7 & 2.7\\
&  Nearest Label Assignment & \textbf{58.9} & \textbf{1.2}\\
\hline
\end{tabular}
\end{center}
    \vspace{-7mm}
\end{table}

\section{Conclusion}
In this paper, we discuss how to design the neural network solution for projection-based LiDAR point cloud segmentation. We propose a pipeline with an input module, a regular backbone, a FID module for upsampling, a classification head, and a post-processing step named NLA. In each module, we first try to choose the most common network settings, then argue why the chosen setting works.

Our pipeline achieves good performance on the benchmark and keeps a simple structure which we believe is hardware-friendly for both GPUs and onboard processors. We released our code, to make further development easier. 

%%%%%%%%%%%%%%%%%%%%%%%%%%%%%%%%%%%%%%%%%%%%%%%%%%%%%%%%%%%%%%%%%%%%%%%%%%%%%%%%

%%%%%%%%%%%%%%%%%%%%%%%%%%%%%%%%%%%%%%%%%%%%%%%%%%%%%%%%%%%%%%%%%%%%%%%%%%%%%%%%

%%%%%%%%%%%%%%%%%%%%%%%%%%%%%%%%%%%%%%%%%%%%%%%%%%%%%%%%%%%%%%%%%%%%%%%%%%%%%%%%

%\section*{APPENDIX}


%\section*{ACKNOWLEDGMENT}

%The preferred spelling of the word ÒacknowledgmentÓ in America is without an ÒeÓ after the ÒgÓ. Avoid the stilted expression, ÒOne of us (R. B. G.) thanks . . .Ó  Instead, try ÒR. B. G. thanksÓ. Put sponsor acknowledgments in the unnumbered footnote on the first page.

%%%%%%%%%%%%%%%%%%%%%%%%%%%%%%%%%%%%%%%%%%%%%%%%%%%%%%%%%%%%%%%%%%%%%%%%%%%%%%%%

%References are important to the reader; therefore, each citation must be complete and correct. If at all possible, references should be commonly available publications.

\bibliographystyle{IEEEtran}
\bibliography{IEEEabrv,IEEEexample}

\onecolumn
\section{Supplementary File with Follow-up Updates}

\subsection{The Goal of the Paper}

Implementing complicated network modules with only one or two points of improvement on hardware is tedious. So here we propose a LiDAR semantic segmentation pipeline on 2D range image just with the most commonly used operators: regular convolutional operator, batch normalization, relu, and bilinear upsample operator. The designed network structure is simple but efficient. We make it achieve comparable performance with state-of-the-art projection-based solutions. The training can be done on a single RTX 2080 Ti GPU.

\subsection{The Change after Official IROS Paper}
\bfsection{The Input Tensor} In the original paper, the input tensor has five channels, x, y, z, range, and remission. We empirically find adding a normal vector for each point will make the training more stable. So we update the input as a eight-channel tensor with x, y, z, range, remission, $n_1$, $n_2$, and $n_3$. $(n_1, n_2, n_3)$ is the normal vector calculated by following \cite{badino2011fast,zhao2021surface}.

\bfsection{The Input Module} The input module in the original paper contains two $1 \times 1$ convolutional layers. We further use five layers to map each point to a higher dimension space.

\bfsection{The Classification Head} We realize the use of the ASPP module is computational heavy. So the updated classification head only contains two $1 \times 1$ convolutional layers with a final softmax layer.

\begin{figure*}[h]
\includegraphics[width=1.0\linewidth,height=45mm]{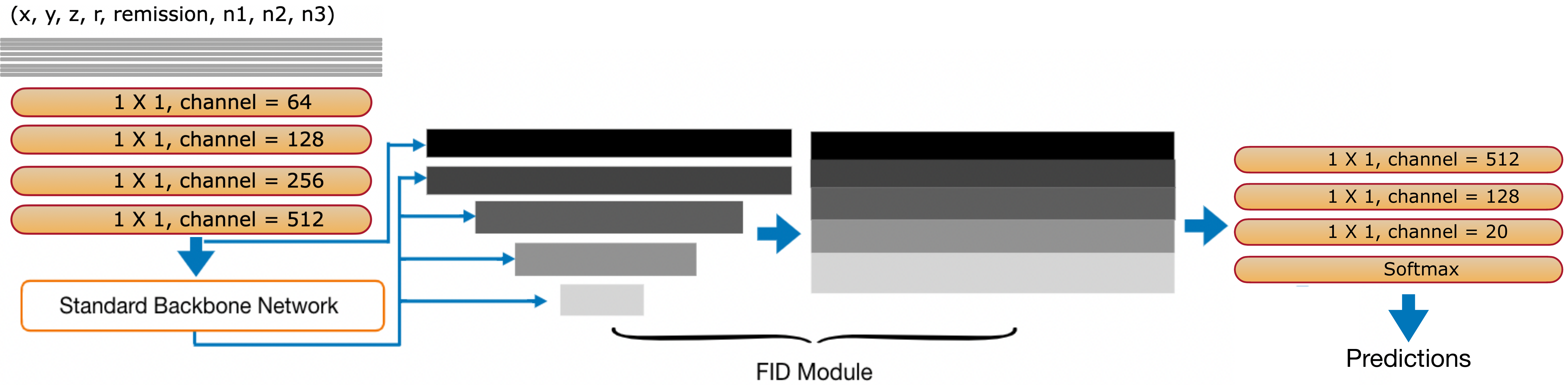}
    \caption{Illustration of our updated network structure. The input module has two $1 \times 1$ layers mapping each point to a high dimensional space. The backbone can be any regular standard network, like ResNet-34 used in this paper. The FID module upsamples all low-resolution feature maps to the original size and concatenates them together. The last classification head takes in the merged large tensor and outputs the label of each point.   }
    \label{fig:new}
    \vspace{-5mm}
\end{figure*}

\bfsection{In Summary} We show the updated network structure in Fig. \ref{fig:new}. The new structure can achieve around 60.0 test mIoU with around 6M parameters in total. Thanks to the half-precision ability of PyTorch, the inference speed is 0.01s for each frame.

\end{document}